\documentclass[10pt,twocolumn,letterpaper]{article}

\usepackage{cvpr}              

\usepackage{graphicx}
\usepackage{amsmath}
\usepackage{amssymb}
\usepackage{booktabs}
\usepackage[dvipsnames]{xcolor}
\makeatletter
\@namedef{ver@everyshi.sty}{}
\makeatother
\usepackage{tikz,pgfplots}
\usepackage{graphicx}
\usepackage{multirow}

\usepackage{bm}
\usepackage{bbm}

\DeclareMathOperator*{\argmin}{arg\,min}

\definecolor{lb}{RGB}{44, 139, 183}
\newcommand\cl[2]{\textcolor{#1}{#2}}

\usepackage[pagebackref,breaklinks,colorlinks]{hyperref}

\usepackage[capitalize]{cleveref}
\crefname{section}{Sec.}{Secs.}
\Crefname{section}{Section}{Sections}
\Crefname{table}{Table}{Tables}
\crefname{table}{Tab.}{Tabs.}

\def\cvprPaperID{9756} 
\def\confName{CVPR}
\def\confYear{2023}

\usepackage{xcolor,colortbl}

\definecolor{Gray}{gray}{0.85}
\definecolor{LightCyan}{rgb}{0.88,1,1}

\newcolumntype{a}{>{\columncolor{Gray}}c}
\newcolumntype{b}{>{\columncolor{white}}c}

\begin{document}

\title{Ranking Regularization for Critical Rare Classes: Minimizing False Positives at a High True Positive Rate}

\author{Kiarash Mohammadi$^{1,2}$\thanks{Work done during an internship at Borealis AI} \quad He Zhao$^1$ \quad Mengyao Zhai$^1$ \quad Frederick Tung$^1$\\
$^{1}$Borealis AI \quad $^{2}$ Mila, Universit{\'e} de Montr{\'e}al\\
{\tt\small kiarash.mohammadi@mila.quebec, \{he.zhao, mengyao.zhai, frederick.tung\}@borealisai.com}
}
\maketitle

\begin{abstract}
In many real-world settings, the critical class is rare and a missed detection carries a disproportionately high cost.
For example, tumors are rare and a false negative diagnosis could have severe consequences on treatment outcomes; fraudulent banking transactions are rare and an undetected occurrence could result in significant losses or legal penalties. In such contexts, systems are often operated at a high true positive rate, which may require tolerating high false positives. 
In this paper, 
we present a novel approach to address the challenge of minimizing false positives for systems that need to operate at a high true positive rate. We propose a ranking-based regularization (\textbf{RankReg}) approach that is easy to implement, and show empirically that it not only effectively reduces false positives, but also complements conventional imbalanced learning losses.
With this novel technique in hand, we conduct a series of experiments on three broadly explored datasets (CIFAR-10\&100 and Melanoma) and show that our approach lifts the previous state-of-the-art performance by notable margins. 
\end{abstract}


\section{Introduction} \label{sec:intro}
\begin{figure}[t]
\begin{center}
   \includegraphics[width=.95\linewidth,trim={30 55 380 0},clip]{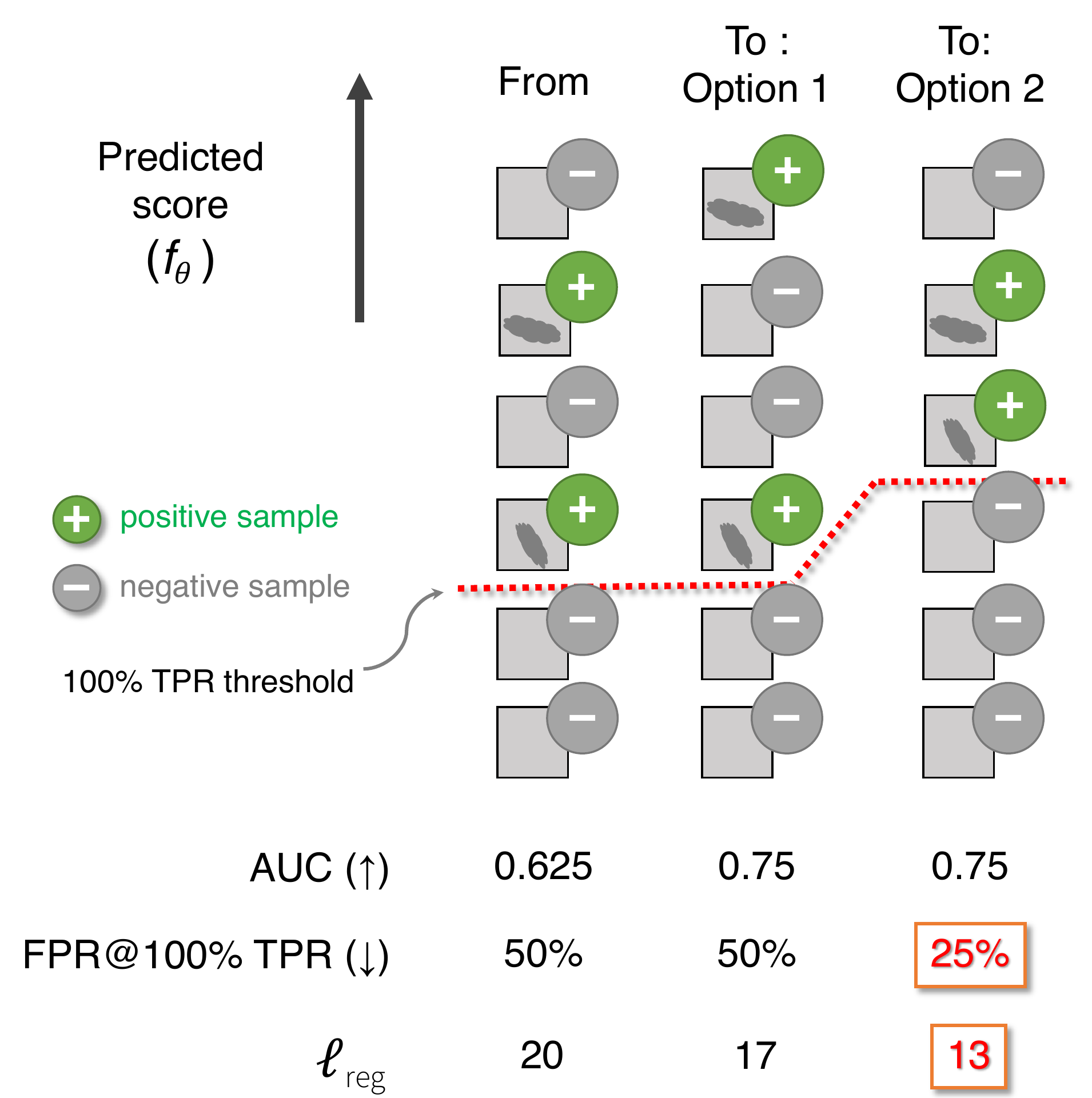}
\end{center}
   \caption{
   Which optimization option is preferred in operational contexts with critical positives?
   We propose a novel regularizer for systems that need to operate at a high true positive rate (TPR). 
   Our approach prioritizes reducing false positives at a high TPR when presented with different options that equally improve the base objective, e.g. area under the ROC curve (AUC). In this toy example, option 2 is preferred because, with a suitable threshold depicted by the dashed line, all positives can be detected (100\% TPR) with only one false positive (\ie, 25\% FPR at 100\% TPR), better than option 1 where two false positives need be tolerated. Our regularizer is consistent with this preference: $\ell_{reg}$ is lower for option 2 than option 1 (\ie, 13 \textit{vs}. 17).
   }
\label{fig:teaser}
\end{figure}

The cost of error is often asymmetric in real-world systems that involve rare classes or events. For example, in medical imaging, incorrectly diagnosing a tumor as benign (a false negative) could lead to cancer being detected later at a more advanced stage, when survival rates are much worse. This would be a higher cost of error than incorrectly diagnosing a benign tumour as potentially cancerous (a false positive). In banking, misclassifying a fraudulent transaction as legitimate may be more costly in terms of financial losses or legal penalties than misclassifying a legitimate transaction as fraudulent (a false positive). In both of these examples, the critical class is rare, and a missed detection carries a disproportionately high cost. In such situations, systems are often operated at high true positive rates, even though this may require tolerating high false positive rates. Unfortunately, false positives can undermine user confidence in the system and responding to them could incur other costs (e.g. additional medical imaging tests).

In this paper, we present a novel approach to address the challenge of minimizing false positives for systems that need to operate at a high true positive rate. Surprisingly, this high-stakes operational setting has rarely been studied by the research community. 
In contrast to conventional imbalanced classification methods,
we propose a general method for inducing a deep neural network to prioritize the reduction of false positives at a high true positive rate. 
To remain as broadly applicable as possible, we make minimal assumptions on the architecture and optimization details of the deep neural network.
Our key insight is that the false positive rate at a high true positive rate is determined by how the least confident positives are ranked by the network.
Our plug-and-play solution adds a simple yet effective ranking-based regularization term to the usual neural network training objective.
The regularization term places an increasing penalty on positive samples the lower they are ranked in a sorted list of the network's classification scores, which works to push up the scores of the hardest positives.

\noindent\textbf{Contributions.} The main contributions of this paper are as follows:
\begin{itemize}
    \item We present a novel plug-and-play regularization term that induces a deep neural network to prioritize the reduction of false positives in operational contexts where a high true positive rate is required.
    \item Our regularizer is generic and can be easily combined with other methods for imbalanced learning.
    \item We conduct extensive experiments on three public benchmarks to show how the proposed regularization term is complementary to conventional imbalanced learning losses, and achieves state-of-the-art performance in the high true positive rate operational setting.
\end{itemize}

\section{Related work}
\noindent\textbf{Imbalanced classification.} 
Class imbalance poses a challenge in training real-world neural network models. Without intervention, the optimization of a classification network becomes dominated by the common classes at the expense of the rare classes. Performing robust and efficient classification on imbalanced data is of practical importance and has seen much recent progress. In general, existing methods for imbalanced classification can be summarized into three major groups: cost function based, model output based, and data based. 
In the first group, various auxiliary optimization objectives are used during training, with a special attention to increase (or balance) the impact of under-represented classes.  One such effort used weighted binary cross entropy (WBCE)~\cite{WBCE}, where the losses of minority samples are multiplied by a scaling factor (typically larger than one) to introduce more cost. Other cost function based approaches include symmetric margin loss (S-ML)~\cite{sml}, symmetric focal loss (S-FL)~\cite{sfl}, asymmetric margin loss (A-ML) and focal loss (A-FL)~\cite{amlafl}, class-balanced BCE (CB-BCE)~\cite{CBBCE} and label distribution aware margin (LDAM)~\cite{LDAM}.
The second group of work tackles the problem by post-processing the model outputs. One simple approach divides the output scores with class co-occurrence frequencies~\cite{IMBsystematic,neuralclassprior}. Similarly, other work adjusts the outputs by re-balancing the probabilities based on minority classes~\cite{Longadjust}. The third group of work focuses on data re-sampling or augmentation. Re-sampling or over-sampling the minority classes has shown solid benefits~\cite{smote2002}. Enhancing samples from rare classes via augmentation also has received much consideration in recent years. Some representative examples are UnMix~\cite{unmix}, ReMix~\cite{remix} and MixUp~\cite{mixup}. 
Though approaching the same problem differently, all aforementioned methods have been used in a variety of vision tasks, \eg, image classification~\cite{smote2002, Byrd2019WhatIT,Han2005BorderlineSMOTEAN,IMBsystematic}, object detection~\cite{chang2021image, li2022equalized} and image/video segmentation~\cite{jadon2020survey,yeung2022unified}, where class imbalance is severe. 

Our work falls into the first group (\ie cost function based). We show in the experiments that our ranking-based regularization term is complementary to a wide range of established cost function based approaches (see Sec.~\ref{sec:exp}). Importantly, in contrast to conventional imbalanced classification methods, we prioritize the reduction of false positives at a high true positive rate. 

\noindent\textbf{Differentiable ranking.} 
Our regularization objective requires ranking
the critical positive samples higher than the negative samples. This task is challenging as ranking operations are piece-wise constant functions, which have zero gradient almost everywhere~\cite{ranksim}. Learning ranking directly through backprop is not feasible. To this end, many alternative solutions have been explored. For instance, an earlier effort considered using the expectation of the ranking as soft-ranker~\cite{softrank}. Some recent work used dynamic programming, instead of backprop, for weights update~\cite{song16icml}, while others achieved this purpose using a differentiable relaxation of histogram binning~\cite{hashingrank,cakiretal2019,revaud19iccv}. Other work has re-formulated it as an optimal
transport problem~\cite{cuturietal2019}. Our work relies on a recent advance that recasts the
ranking operation as the minimizer of a linear combinatorial objective~\cite{ranksolver}, which can be sovled elegantly by a blackbox
combinatorial solvers~\cite{blackboxsolver} that can provide informative gradients from a continuous interpolation.

\noindent\textbf{Deep AUC optimization.} Our work can be considered (and will be demonstrated in Section ~\ref{sec:exp}) as an optimizer for maximizing the AUC score, \textit{cf.}~\cite{auc1,auc3,auc2,auc4}. Nonetheless, none of them are designed to favor low false positive rates 
given a high true positive rate requirement.
The work most similarly motivated to ours is ALM~\cite{alm}, which recently advocated yielding higher score (\ie probability) for critical positives than non-critical negatives, thus improving the AUC. ALM formulates the problem from a constrained optimization perspective and achieves strong empirical results. We provide comprehensive comparisons with this state-of-the-art baseline in our experiments.
\section{Technical approach}
We present a novel, plug-and-play regularization loss as a generic method for inducing a neural network to prioritize minimizing false positives at a high true positive rate.
First, we formulate the imbalanced binary classification task with critical positives in Sec.~\ref{sec:problem-formulation}. Second, to obtain a solution that is tailored asymmetrically to the high true positive rate setting, we introduce a ranking-based regularization term that encourages models to rank the critical positives higher than the non-critical negatives, 
while prioritizing reducing the false positive rate at high true positive rate thresholds.
Third, we discuss the used optimization solution in Sec.~\ref{sec:backprop}, as the ranking operation is challenging to optimize with back-propagation due to its non-differentiable nature. 

\begin{figure}[t]
\begin{center}
   \includegraphics[width=\linewidth,trim={0 30 240 0},clip]{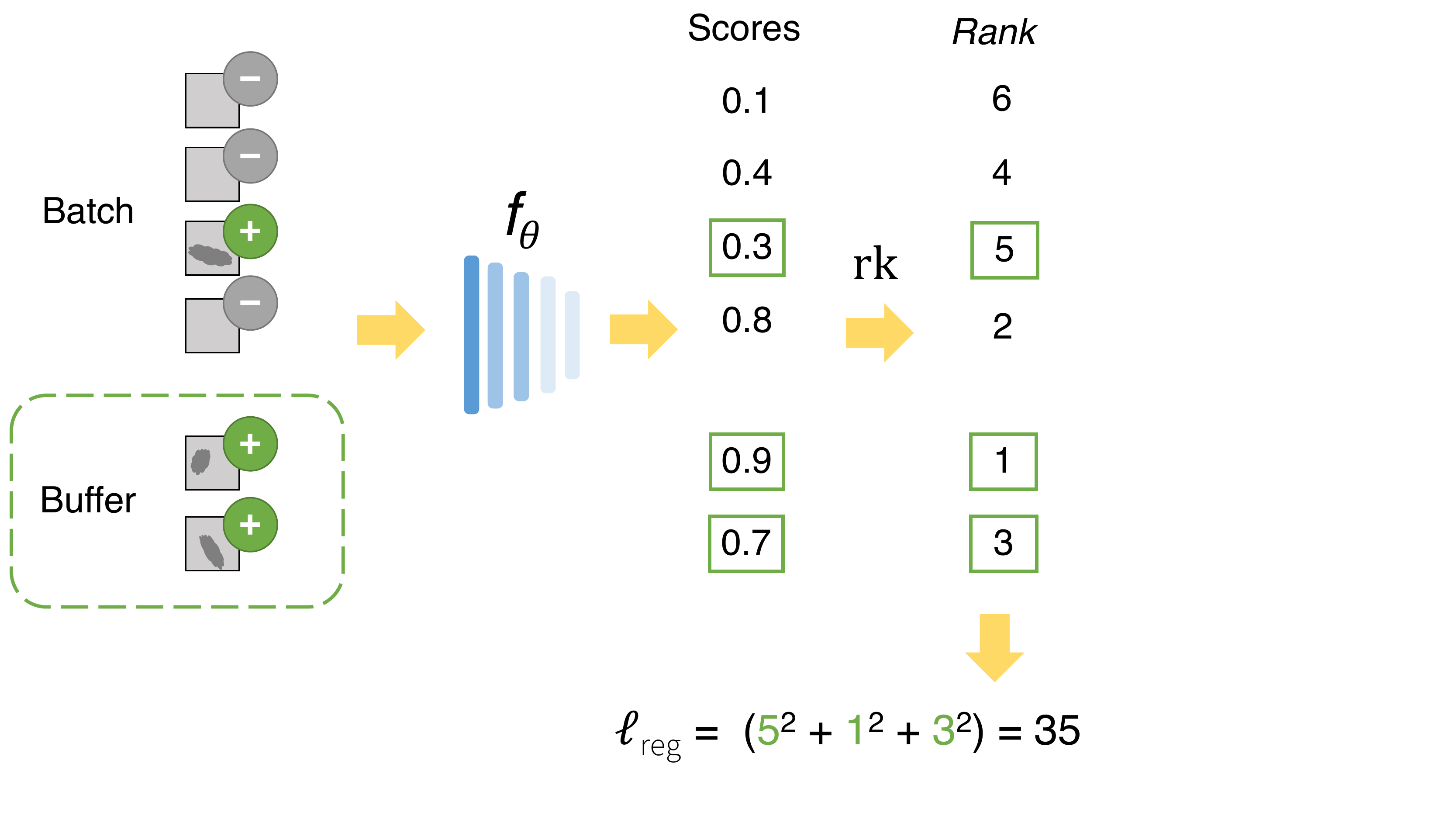}
\end{center}
    \caption{An illustration of the regularization term by example. Given a batch of inputs and a buffer (\ie, external memory) of positive samples, we collect the probability scores from the classifier, $f_\theta$, and apply the ranking function, $\bm{rk}(\cdot)$, to obtain a vector $\bm{r}$ of rank values. The rank values associated with the positive samples (\ie, numbers in \textcolor{LimeGreen}{green} boxes) are then used to compute the ranking-based regularization loss, $\ell_{reg}$, using Eq.~\ref{eq:ell_reg}; we drop the normalization in this figure for simplicity of presentation. Note that $f_{\theta}$ can be an arbitrary classification architecture and RankReg does not rely on the specifics of that approach.}
\label{fig:method}
\end{figure}

\subsection{Problem formulation}
\label{sec:problem-formulation}
We seek to predict the correct label over a highly imbalanced binary classification dataset $D = \{(\bm{x}_1, y_1), (\bm{x}_2, y_2), \dots, (\bm{x}_n, y_n), y_i \in \{0, 1\}\}$, where $\bm{x}_i$ are data samples, $y_i$ are labels, and the critical data samples (positive class, labelled 1) appear much less frequently than the non-critical data samples (negative class, labelled 0). For example, medical images with cancerous tumors may be critical positives, while images with benign tumors or no tumors may be negatives. It is assumed that the cost of missing the positive class (a false negative) carries a disproportionately high cost, and that the system is required to operate at a high true positive rate. 

Our goal is to produce a general method for inducing a deep neural network (DNN) classifier $f_{\theta}: \mathbb{R}^{d} \to \mathbb{R}$, mapping $d$-dimensional inputs to output scores, to prioritize the reduction of false positives at a high true positive rate~\cite{alm}. To be as general as possible, the method should make minimal assumptions on the architecture and optimization details of $f_{\theta}$.

\subsection{Preliminary}

We propose a novel ranking-based~\cite{ranksolver, ranksim} regularizer that fulfills the above desiderata.
Our key insight is that the false positive rate at a high true positive rate is determined by how the least confident positives are ranked by the network.
Our plug-and-play approach adds a regularization term to the usual DNN training objective, making our solution complementary to a wide range of base objective functions, from conventional binary cross-entropy to more sophisticated imbalanced losses such as asymmetric focal loss~\cite{amlafl}. In other words, we modify the DNN training objective to be:
\begin{equation}
    \ell(f_{\theta}(\bm{x}), \bm{y}) = \ell_{base}(f_{\theta}(\bm{x}), \bm{y}) + \lambda \; \ell_{reg}(f_{\theta}(\bm{x}), \bm{y}),
\end{equation}
where $\ell_{base}$ is the base objective function, $\ell_{reg}$ is the new regularization term, and $\lambda$ is a balancing hyperparameter with value empirically set to 1.

\subsection{Ranking regularizer}~\label{sec:RankReg}



Denote by $\bm{rk}(\cdot)$ the ranking function that takes a vector of real values $\bm{a}$ and outputs the rank of each element in the sorted vector. In other words, the $i$th element in $\bm{rk}(\bm{a})$ is given by 
\begin{equation}
\bm{rk}(\bm{a})_i = 1 + |\{j: \bm{a}_j > \bm{a}_i\}|.
\label{eq:rk}
\end{equation}

Then, we devise a regularization term that is computed as the normalized sum over the squared rank values of the positive samples:
\begin{align}
    \bm{r} = \bm{rk} \left( \left[f_{\theta}(\bm{x}_1), f_{\theta}(\bm{x}_2), \dots, f_{\theta}(\bm{x}_B) \right] \right), \\
    \ell_{reg}(f_{\theta}(\bm{x}), \bm{y}) = \frac{1}{|P|} \sum_{i=1}^{B} \bm{r}_i^{2} \cdot \mathbbm{1}[y_i = 1],
    \label{eq:ell_reg}
\end{align}
where $|P| = \sum_{i=1}^{B} \mathbbm{1}[y_i = 1]$ is the number of positive samples in the batch of $B$ samples. In practice, since positive samples may be severely under-represented in the dataset, we compute the regularization term over the union of the batch and an external memory that caches previous positive samples; we will revisit the implementation details of this buffer in Section \ref{sec:buffer}. We also normalize the rank values $\bm{r}$ to be between 0 and 1.



To see how this regularization term prioritizes the reduction of false positives at a high true positive rate, let us consider the toy example in Figure~\ref{fig:teaser}. Suppose that the classifier $f_\theta$ currently produces the sorted ordering shown in the left column. The ground-truth critical positives are represented by green plus icons and the ground-truth negatives are represented by grey minus icons. The positive examples induce the second and fourth highest classification scores (higher is better for positives). To achieve a high true positive rate of 100\% on these two positives, we would have to accept at least two false positives, obtaining a false positive rate of 50\%. Now, suppose that in the next training iteration, the optimizer has two options, shown in the middle and right columns, that would equally improve the training objective; here, we illustrate with the area under the ROC curve (AUC), a common retrieval-based objective. While equally preferable by the training objective, the right column is better aligned with our goal of reducing false positives at a high true positive rate: with a suitable threshold (depicted by the dashed line), we can obtain a false positive rate of 33\% at a true positive rate of 100\%. On the other hand, the middle column can at best achieve a false positive rate of 50\% at a true positive rate of 100\%.

The proposed regularization term distinguishes between the middle and right columns, and assigns a higher loss to the middle column. In the middle column, the positives have the first and fourth highest classification scores, producing a regularization loss of $1^2 + 4^2 = 17$ (we drop the normalization terms here for simplicity of presentation). In the right column, the positives have the second and third highest classification scores, producing a regularization loss of $2^2 + 3^2 = 13$. The proposed regularization therefore favors the right column, as desired. Note that if we use the ranks directly instead of squaring, the regularization loss would be $5$ in both cases ($1+4=2+3=5$).
Squaring places an increasing penalty on positive samples the lower they are ranked in a sorted list of the network's output scores, which works to push up the scores of the least confident positive samples.

%

\subsection{Optimization on ranking operations}
\label{sec:backprop}

Rank-based objectives often arise in computer vision \cite{softrank, directloss, hashingrank} 
and are typically challenging to optimize due to the non-differentiability of the ranking function.
The ranking function is piece-wise constant, i.e., perturbing the input would most likely not change the output. Thus, we cannot obtain informative gradients (i.e., gradients are zero almost everywhere).  
We adopt the optimization approach of \cite{rolineketal2020}, which frames the ranking function as a combinatorial solver and relies on an elegant way of backpropagating through blackbox combinatorial solvers \cite{ranksolver}. The combinatorial objective version of computing the ranking function is given by
\begin{equation}
    \bm{rk}(\bm{a}) = \argmin_{\bm{\pi} \in \Pi_n} \bm{a} \cdot \bm{\pi},
\end{equation}
for an arbitrary vector $\bm{a}$ of real values (see Eq.~\ref{eq:rk}), where $\Pi_n$ is a set that contains all the permutations of $\{1, 2, \cdots, n\}$. 
This reframing enables us to leverage \cite{ranksolver} to differentiate through a blackbox combinatorial solver: \cite{ranksolver} proposes a family of piecewise affine continuous
interpolation functions parameterized by a single hyperparameter that controls the tradeoff between faithfulness to the true function and informativeness of the gradient.
In brief, we compute and return the gradient of the continuous interpolation:
$    \frac{\partial \mathcal{L}}{\partial \mathbf{a}} = -\frac{1}{\gamma} (\mathbf{rk}(\mathbf{a}) - \mathbf{rk}(\mathbf{a}_\gamma))$,
where $\mathbf{a}_\gamma$ is a perturbed input derived from the incoming gradient information $\frac{\partial \mathcal{L}}{\partial \mathbf{rk}}$ via
$\mathbf{a}_\gamma = \mathbf{a} + \gamma \, \frac{\partial \mathcal{L}}{\partial \mathbf{rk}}$.
We refer the interested reader to \cite{rolineketal2020} for further details.

\subsection{Buffer of positive samples}
\label{sec:buffer}

%
During training, 
we maintain a buffer (\ie, an external memory) of positive samples to enable the regularization term to be computed per batch 
even in datasets with severe imbalance ratios, as a batch may contain few (or no) positive samples.
The whole buffer is always appended to the batch.
We implement the buffer as a priority queue and it works as follows.
At the start of training, positive samples are accumulated from the incoming batches and added to the buffer up to a fixed maximum capacity. Afterwards, as batches are processed, new positive samples replace the samples in the buffer for which the model is the most certain, \ie, the buffered samples with the maximum $f_\theta$ responses. This replacement strategy keeps the hard positives in the buffer and removes positives for which the classifier is already confident. We consider the alternative replacement strategies of first-in-first-out (FIFO) and minimum $f_\theta$ responses
in the ablation studies.
The complete pipeline, including the buffer of positive samples, is illustrated by example in Figure~\ref{fig:method} (without normalization for simplicity of presentation).


%
\section{Empirical evaluation}
\label{sec:exp}

\subsection{Overview}
To demonstrate the effectiveness of RankReg, extensive experiments are conducted on three public image-based benchmarks: binary imbalanced CIFAR-10, binary imbalanced CIFAR-100, and Melanoma. To adapt CIFAR-10 and CIFAR-100 to the critical positives setting, we follow the same experimental protocol as the state-of-the-art baseline \cite{alm}. As we do not have access to the private medical imaging dataset used in \cite{alm}, we performed experiments on a publicly available medical imaging dataset (\ie, Melanoma). In this section, we first describe the benchmarks, baselines, and training details. We then present experimental results on the three benchmarks, showing how the proposed regularizer is complementary to conventional imbalanced learning losses and achieves state-of-the-art results. We conclude with ablations and analyses of design choices. 

\begin{table}[t]
\small
\centering 
\begin{tabular}{l c c c c} 
\toprule 
& \multicolumn{4}{c}{Binary CIFAR10, imb. \textbf{1:100}} \\
\cmidrule(l){2-5} 
Methods & \vtop{\hbox{\strut FPR@ $\downarrow$}\hbox{\strut  98\%TPR}} & \vtop{\hbox{\strut FPR@ $\downarrow$}\hbox{\strut  95\%TPR}} &\vtop{\hbox{\strut FPR@ $\downarrow$}\hbox{\strut  92\%TPR}} &  AUC $\uparrow$ \\ 
\midrule 
BCE & 56.0 & 45.0 & 29.0 &  91.2 \\ 
+ALM  & 52.0 & 34.0 & 21.0 & 93.1\\ 
+RankReg   & \underline{\textbf{47.1}} & \underline{\textbf{26.2}} & \underline{\textbf{20.6}} & \textbf{\cl{red!60!white}{94.3}} \\ 
\midrule
S-ML & 59.0 & 40.0 & 26.0 &  91.7 \\ 
+ALM  & 50.0 & 37.0 & \underline{\textbf{24.0}} &  92.5\\ 
+RankReg  & \underline{\textbf{45.6}} & \underline{\textbf{31.4}} & 29.7 &  \textbf{\cl{red!60!white}{93.9}} \\ 
\midrule
S-FL & 59.0 & 40.0 & 27.0 &  91.7 \\ 
+ALM  & 55.0 & 39.0 & 25.0 &  91.5\\ 
+RankReg   & \underline{\textbf{53.3}} & \underline{\textbf{35.4}} & \underline{\textbf{20.7}}  & \textbf{\cl{red!60!white}{92.8}} \\ 
\midrule
A-ML & 54.0 & 36.0 & 23.0 &  92.4 \\ 
+ALM & \underline{\textbf{45.0}} & 35.0 & 23.0 &  92.8\\ 
+RankReg  & 47.8 & \underline{\textbf{28.9}} & \underline{\textbf{21.4}} &  \textbf{\cl{red!60!white}{94.1}} \\ 
\midrule
A-FL & 50.0 & 38.0 & 24.0 &  92.3 \\ 
+ALM & \underline{\textbf{49.0}} & 37.0 & 23.0 &  92.8\\ 
+RankReg  & 50.5 & \underline{\textbf{28.7}} & \underline{\textbf{20.9}} &  \textbf{\cl{red!60!white}{94.3}} \\ 
\midrule
CB-BCE & 89.0 & 72.0 & 59.0 &  78.0 \\ 
+ALM  & 67.0 & 51.0 & 36.0 &  88.1\\ 
+RankReg   & \underline{\textbf{48.8}} & \underline{\textbf{29.9}} & \underline{\textbf{24.6}} & \textbf{\cl{red!60!white}{93.2}} \\ 
\midrule
W-BCE & 69.0 & 52.0 & 37.0 &  87.4 \\ 
+ALM  & 66.0 & 48.0 & 31.0 &  89.3\\ 
+RankReg  & \underline{\textbf{60.0}} & \underline{\textbf{39.4}} & \underline{\textbf{29.6}} &  \textbf{\cl{red!60!white}{92.1}} \\ 
\midrule
LDAM & 65.0 & 48.0 & 34.0 &  89.0 \\ 
+ALM  & 60.0 & 42.0 & 31.0 &  91.0\\ 
+RankReg  & \underline{\textbf{42.8}} & \underline{\textbf{25.6}} & \underline{\textbf{23.8}} &  \textbf{\cl{red!60!white}{95.0}} \\ 
\midrule
\textcolor{Green}{Avg.} $\Delta$ & \textcolor{Green}{6.0}  & \textcolor{Green}{9.7} & \textcolor{Green}{2.8} & \textcolor{Green}{2.3} \\
\bottomrule 

\end{tabular}
\caption{
Comparison results for binary imbalanced CIFAR-10 showing FPRs at \{98\%, 95\%, 92\%\} TPRs. Baseline numbers are quoted from ALM~\cite{alm}. ``+ALM" and ``+RankReg" are shorthand for \textit{BaseLoss}+ALM and \textit{BaseLoss}+RankReg, respectively.
} 
\label{tab:cifar10_imb100} 
\end{table}

\begin{table}[t]
\small
\centering 
\begin{tabular}{l c c c c} 
\toprule 
& \multicolumn{4}{c}{Binary CIFAR100, imb. \textbf{1:100}} \\ %
\cmidrule(l){2-5} 
Methods & \vtop{\hbox{\strut FPR@ $\downarrow$}\hbox{\strut  98\%TPR}} & \vtop{\hbox{\strut FPR@ $\downarrow$}\hbox{\strut  95\%TPR}} &  \vtop{\hbox{\strut FPR@ $\downarrow$}\hbox{\strut  90\%TPR}} & AUC $\uparrow$ \\ 
\midrule 
BCE & 93.0 & 63.0 & 47.0 & 81.8 \\ 
+ALM  & 91.0 & 49.0  & 39.0 & 82.7\\ 
+RankReg   & \underline{\textbf{85.2}} & \underline{\textbf{42.4}}  & \underline{\textbf{28.7}} & \textbf{\cl{red!60!white}{85.5}} \\ 
\midrule
S-ML & 89.0 & 65.0  & 43.0 & 82.7 \\ 
+ALM  & 88.0 & 69.0  & 41.0 & 81.7\\ 
+RankReg  & \underline{\textbf{64.0}} & \underline{\textbf{44.8}}  & \underline{\textbf{34.5}} & \textbf{\cl{red!60!white}{85.4}} \\ 
\midrule
S-FL & 89.0 & 62.0  & 44.0 & 82.6 \\ 
+ALM  & 88.0 & 60.0  & 42.0 & 81.7\\ 
+RankReg   & \underline{\textbf{84.6}} & \underline{\textbf{49.2}} & \underline{\textbf{38.4}} & \textbf{\cl{red!60!white}{84.7}} \\ 
\midrule
A-ML & 91.0 & 63.0  & 44.0 & 81.8 \\ 
+ALM & 89.0 & 55.0  & 37.0 & 82.7\\ 
+RankReg  & \underline{\textbf{81.6}} & \underline{\textbf{43.4}}  & \underline{\textbf{32.6}} & \textbf{\cl{red!60!white}{85.5}} \\ 
\midrule
A-FL & 88.0 & 63.0  & 45.0 & 82.8 \\ 
+ALM & 86.0 & 62.0  & 40.0 & 83.2\\ 
+RankReg  & \underline{\textbf{70.0}} & \underline{\textbf{53.4}}  & \underline{\textbf{35.8}} & \textbf{\cl{red!60!white}{84.6}} \\ 
\midrule
CB-BCE & 93.0 & 75.0  & 52.0 & 78.8 \\ 
+ALM  & \underline{\textbf{89.0}} & 59.0  & 36.0 & 83.8 \\ 
+RankReg   & 89.8 & \underline{\textbf{48.6}} & \underline{\textbf{33.4}} & \textbf{\cl{red!60!white}{84.1}} \\ 
\midrule
W-BCE & 88.0 & 59.0  & 41.0 & 79.7 \\ 
+ALM  & 87.0 & \underline{\textbf{53.0}}  & \underline{\textbf{39.0}} & \textbf{\cl{red!60!white}{83.2}} \\ 
+RankReg  & \underline{\textbf{84.0}} & 60.0  & 41.1 & 82.9 \\ 
\midrule
LDAM & 84.0 & 70.0  & 42.0 & 82.3 \\ 
+ALM  & 80.0 & 59.0  & 40.0 & 83.2\\ 
+RankReg  & \underline{\textbf{70.3}} & \underline{\textbf{51.6}}  & \underline{\textbf{35.0}} & \textbf{\cl{red!60!white}{84.7}} \\ 
\midrule
\textcolor{green!60!black}{Avg.} $\Delta$ & \textcolor{Green}{8.6} & \textcolor{Green}{8.6} & \textcolor{Green}{4.3} & \textcolor{Green}{1.9} \\
\bottomrule 
\end{tabular}
\caption{
Comparison results for binary imbalanced CIFAR-100 showing FPRs at \{98\%, 95\%, 90\%\} TPRs. Baseline numbers are quoted from ALM~\cite{alm}. ``+ALM" and ``+RankReg" are shorthand for \textit{BaseLoss}+ALM and \textit{BaseLoss}+RankReg, respectively.
} 
\label{tab:cifar100_imb100} 
\end{table}

\subsection{Datasets and evaluation protocol}
\label{sec:dataset_eval}
\noindent\textbf{Binary imbalanced CIFARs.} 
For a fair comparison, we adopt the same binary imbalanced versions of CIFAR-10 and 100 as curated by the authors of the state-of-the-art method~\cite{alm}. In brief, binary imbalanced CIFAR-10 is constructed by randomly designating two classes as positives and negatives. All training samples from the negative class are used, while training samples from the positive class are subsampled. Binary imbalanced CIFAR-100 is constructed by designating one super-class as the negative class and a sub-class of a different super-class as the positive class. Again, all training samples from the negative class are used, while training samples from the positive class are subsampled. We refer the interested reader to \cite{alm} for the construction details. Following the evaluation protocol in \cite{alm}, we experiment with 1:100 and 1:200 imbalance ratios (1 critical positive to 100 or 200 negatives), set aside 100 and 50 samples per class to form the validation set for hyper-parameter selection for CIFAR-10 and 100 respectively, and evaluate on a class-balanced test set.  

\noindent\textbf{Melanoma.} The Kaggle Melanoma dataset is a medical image classification dataset that was first proposed on a competition for identifying melanoma (a common form of skin cancer) in imaging scans of skin lesion~\cite{melanoma}. It is composed of 33,126 images collected from patients in large variance, where only 584 out of the entire set are malignant (\ie, positive) melanoma; therefore, the dataset has a 1:176 imbalance ratio. It is split into training, validation, and test sets with ratios of 70\%, 10\%, and 20\%, respectively. The original resolutions of
images in Melanoma are too high (\eg, [6000, 4000] or [1920, 1080]) to fit in the backbone network (\ie, ResNet-18). We resize them into [256, 256] for the convenience of computational resources, \textit{cf.}~\cite{auc1}. Since Melanoma is naturally imbalanced, no further curation is needed for our study.

\noindent\textbf{Metrics.}
We evaluate the performance using the false positive rate (FPR) against several increasingly strict true positive rates (TPR), \ie, FPR@\{$\mathbf{\beta}$\}TPR and $\beta \in \{90\%, 92\%, 95\%, 98\% \}$. For completeness, we also evaluate using the area under curve metric (AUC) to reveal the overall classification performance, as typically seen in related work~\cite{auc1,auc2,auc3,auc4}. 

\noindent\textbf{Baseline methods.}
\label{sec:base_objective_functions} Following previous method~\cite{alm}, we consider applying our proposed regularizer with several different existing loss functions, most of which have been designed to handle class imbalance: binary cross-entropy (BCE), symmetric margin loss (S-ML)~\cite{sml}, symmetric focal loss (S-FL)~\cite{sfl}, asymmetric margin loss (A-ML) and focal loss (A-FL)~\cite{amlafl}, cost-weighted BCE (WBCE)~\cite{WBCE}, class-balanced BCE (CB-BCE)~\cite{CBBCE}, and label distribution aware margin (LDAM)~\cite{LDAM}.

\subsection{Implementation details}
\noindent\textbf{Backbone architectures.} For the binary imbalanced CIFAR datasets, we adopt the ResNet-10 architecture up to the second last layer as feature extractor and append a multi-layer perceptron (MLP) with shape [512$\rightarrow$ 2] as the classifier. For Melanoma, 
we adopt the richer architecture of ResNet-18 and repeat the same step to create the classifier. 

\noindent\textbf{Buffer usage.} The buffer of positive samples can have a rebalancing effect when used to compute the base loss in addition to the usual batch samples. We leverage this effect when training models on the binary imbalanced CIFAR-10 and CIFAR-100, which have balanced test sets following the protocol in \cite{alm}. Since all splits in the Melanoma dataset follow the data's natural class imbalance, when training the Melanoma models we compute the base loss using the batch samples only. Throughout our experiments, we set the batch size to 64 when training other methods (including base methods and ALM). 
When training RankReg models, we use a buffer size of 32 and reduce the batch size to 32. 

\noindent\textbf{Hyper-parameter search.} For CIFAR-10 and CIFAR-100, which we have in common with~\cite{alm}, we use the same hyper-parameters on all base loss functions, for the sake of fair comparison.
For Melanoma dataset, we tune the hyper-parameters as follows. The more general parameters like learning rate and batch size are chosen and fixed to work with the BCE loss. For ALM \cite{alm}, a two step grid-search is performed. In the first step, we perform a grid-search over $\rho$ and $\mu^{(0)}$. We choose $\rho$ from the set $\{2, 3\}$ and $\mu^{(0)}$ from the set $\{10^{-2}, 10^{-3}, 10^{-4}, 10^{-5}, 10^{-6}, 10^{-7}, 10^{-8}, 10^{-9}\}$ (note that this is a slightly more thorough grid-search than the original paper). When these two are fixed, we search for the best $\delta$ from the set $\{0.1, 0.25, 0.5, 1.0\}$. These parameters are tuned based on the AUC on the validation set. 



\noindent\textbf{Model ensembling.} For CIFAR-10 and CIFAR-100 where the datasets are rather small, we report results from 10 ensembling models for higher reliability and to diminish dataset-dependant biases, matching the protocol in \cite{alm}. In detail, 10 random stratified splits of the dataset are created and a model is trained on each. Finally, these models are ensembled by averaging their outputs in the logit space. 
We do not perform ensembling on Melanoma, as it is larger and has a standard data split.

\begin{table}[t]
\centering 
\resizebox{\columnwidth}{!}{%
\begin{tabular}{l c c c c c} 
\toprule 
& \multicolumn{5}{c}{Melanoma, imb. \textbf{1:170}} \\ %
\cmidrule(l){2-6} 
Methods & \vtop{\hbox{\strut FPR@ $\downarrow$}\hbox{\strut  98\%TPR}} & \vtop{\hbox{\strut FPR@ $\downarrow$}\hbox{\strut  95\%TPR}} &\vtop{\hbox{\strut FPR@ $\downarrow$}\hbox{\strut  92\%TPR}} &  \vtop{\hbox{\strut FPR@ $\downarrow$}\hbox{\strut  90\%TPR}} & AUC $\uparrow$ \\ 
\midrule 
BCE & 49.8 & 45.9 & 38.6 & 35.5 & 85.7 \\ 
+ALM  & 49.9 & 41.8 & 40.0 & 37.7 & 85.6\\ 
+RankReg  & \underline{\textbf{49.4}} & \underline{\textbf{37.9}} & \underline{\textbf{33.9}} & \underline{\textbf{31.6}} & \textbf{\cl{red!60!white}{86.8}} \\ 
\midrule
S-ML & \underline{\textbf{46.6}} & 42.8 & 38.4 & 37.4 & 85.3 \\ 
+ALM  & 51.3 & \underline{\textbf{40.5}} & 39.8 & 36.2 & 83.5\\ 
+RankReg  & 54.6 & 42.4 & \underline{\textbf{36.1}} & \underline{\textbf{34.4}} & \textbf{\cl{red!60!white}{86.3}} \\ 
\midrule
S-FL & 59.0 & 47.3 & 44.4 & 39.5 & 83.8 \\ 
+ALM  & \underline{\textbf{47.8}} & 42.7 & 39.2 & 38.1 & 84.0\\ 
+RankReg   & 56.6 & \underline{\textbf{37.8}} & \underline{\textbf{31.2}} & \underline{\textbf{29.8}} & \textbf{\cl{red!60!white}{86.1}} \\ 
\midrule
A-ML & \underline{\textbf{47.5}} & 42.9 & 40.4 & 36.6 & 85.4 \\ 
+ALM & 51.0 & 41.5 & 37.5 & 37.1 & 83.7 \\ 
+RankReg  & 58.3 & \underline{\textbf{40.8}} & \underline{\textbf{36.7}} & \underline{\textbf{33.9}} & \textbf{\cl{red!60!white}{86.2}} \\ 
\midrule
A-FL & 55.6 & 45.0 & 42.7 & 41.2 & 84.4 \\ 
+ALM & 49.0 & 42.4 & 40.1 & 38.1 & 83.6\\ 
+RankReg  & \underline{\textbf{48.0}} & \underline{\textbf{36.2}} & \underline{\textbf{30.7}} & \underline{\textbf{28.8}} & \textbf{\cl{red!60!white}{86.3}} \\ 
\midrule
CB-BCE & 67.2 & 59.5 & 35.7 & \underline{\textbf{33.2}} & 82.6 \\ 
+ALM  & 60.8 & 59.5 & 46.3 & 45.8 & 81.5\\ 
+RankReg & \underline{\textbf{57.8}} & \underline{\textbf{44.9}} & 35.7 & 34.7 & \textbf{\cl{red!60!white}{83.7}} \\ %
\midrule 
W-BCE & 69.0 & 52.0 & 37.0 & 32.1 & 87.4 \\ 
+ALM  & 66.0 & 48.0 & \underline{\textbf{31.0}} & 30.7 & 89.3\\ 
+RankReg  & \underline{\textbf{56.4}} & \underline{\textbf{41.1}} & 33.0 & \underline{\textbf{30.5}} & \textbf{\cl{red!60!white}{90.9}} \\ 
\midrule
LDAM & \underline{\textbf{59.7}} & 48.2 & 46.2 & \underline{\textbf{39.0}} & \textbf{\cl{red!60!white}{83.4}} \\ 
+ALM  & 62.7 & 47.7 & \underline{\textbf{43.3}} & 40.7 & 81.5\\ 
+RankReg  & 65.6 & \underline{\textbf{47.5}} & 45.7 & 43.9 & 81.7 \\ 
\bottomrule 
\end{tabular}
}
\caption{ Comparison results for Melanoma dataset showing FPRs at \{98\%, 95\%, 92\%, 90\%\} TPRs. ``+ALM" and ``+RankReg" are shorthand for \textit{BaseLoss}+ALM and \textit{BaseLoss}+RankReg, respectively.
} 
\label{tab:melanoma} 
\end{table}

\subsection{Comparison to alternative approaches}
\label{sec:main_rst}
\noindent\textbf{Binary imbalanced CIFAR-10.} Table~\ref{tab:cifar10_imb100} compares the performance of using RankReg as well as the previous state-of-the-art method ALM~\cite{alm} together with eight base losses on CIFAR-10 dataset, curated with a imbalance ratio of 1:100 (see Sec.~\ref{sec:dataset_eval}). We group the empirical results by base loss (BCE, S-ML, S-FL, etc.).  Within each group, we first show the results obtained by applying the base method as well as the previous state-of-the-art approach. Then, we show our results. For each FPR and overall AUC, the best result is either underlined or highlighted in red text.

It is clear that our results are consistently better on most metrics, except for three FPR values at S-ML, A-FL and A-ML baselines, where RankReg is the second best approach. The performance improvement is especially striking when coupling RankReg with CB-BCE: RangReg reduces the FPR at the strictest TPR ratio by 18\%, \ie, from 67.0 to 48.8 in FPR@98\%TPR. The best overall results are obtained by fusing RankReg with the LDAM baseline, where we achieve the higest AUC score (\ie, 95.0) as well as the lowest FPR@98\%TPR value (\ie, 42.8) across all experimental results. 

Even though our goal is \textit{not} to have higher AUC scores, our approach obtains the new state-of-the-art AUC performance on all baselines. 
The bottom row in Table~\ref{tab:cifar10_imb100} shows the improvements on FPRs using RankReg compared to previous best results, averaged across all baselines. It is notable that our approach obtains larger gains at higher TPRs. For instance, our approach achieves 6.0 and 9.7 FPR improvements at 98\% and 95\% TPRs, respectively, compared to 2.8 at 92\% TPR, which is  favored by our goal (see Sec.~\ref{sec:RankReg}).

\noindent\textbf{Binary imbalanced CIFAR-100.} 
To show the capability of our method to scale, we evaluate our method on the curated CIFAR-100 dataset. The results in Table~\ref{tab:cifar100_imb100} are consistent with our results on CIFAR-10. Once again, our approach is the top performer across most metrics. 
However, this time, both the BCE and A-ML baselines achieve the highest AUC score using RankReg. Moreover, it is notable that on the highest TPR (\ie, 98\%), our approach outperforms the previous state-of-the-art with the margin $>$ 10\% on 5 different baselines (\ie, A-FL, S-ML, BCE, LDAM and S-FL). Such notable gains are only observed twice in previous experiments (\ie, LDAM and CB-BCE in Table~\ref{tab:cifar10_imb100}).

\begin{table}[]
\centering
\resizebox{\columnwidth}{!}{%
\begin{tabular}{@{}lllllcllll@{}}\toprule[1.2pt]
\multicolumn{1}{c}{\multirow{ 2}{*}{FPR@$\beta$TPR}} & \multicolumn{4}{c}{\textbf{CIFAR10}} & \phantom{a}& \multicolumn{4}{c}{\textbf{CIFAR100}} \\
\cmidrule{2-5} \cmidrule{7-10}
    & 98\%   & 95\%  & 92\% & AUC   && 98\%   & 95\%   & 90\% & AUC \\
 \midrule
Ranks & 52.1 & 35.2 & 24.0 & 93.6 && 86.3 & 52.8 & 43.0 & 83.2 \\
Squared ranks & 47.1 & 26.2 & 20.6 & 94.3 && 85.2 & 42.4 & 28.7 & 85.5 \\
Cubed ranks & 45.5 & 31.9 & 23.0 & 93.7 && 84.2 & 53.8 & 50.4 & 83.7 \\
Exponential of ranks & 44.5 & 34.0 & 24.3 & 93.6 && 83.6 & 48.8 & 39.4 & 84.9 \\
\bottomrule[1.2pt]
\end{tabular}
}
\caption{Ablation study of different ranking penalty choices on imbalanced CIFAR-10 and 100 datasets.}
\label{tab:rankpenalty}
\end{table}



\begin{table}[]
\centering
\resizebox{\columnwidth}{!}{%
\begin{tabular}{@{}lllllcllll@{}}\toprule[1.2pt]
\multicolumn{1}{c}{\multirow{ 2}{*}{FPR@$\beta$TPR}} & \multicolumn{4}{c}{\textbf{CIFAR100}} & \phantom{a}& \multicolumn{4}{c}{\textbf{Melanoma}} \\
\cmidrule{2-5} \cmidrule{7-10}
    & 98\%   & 95\%  & 92\% & AUC   && 98\%   & 95\%   & 92\% & AUC \\
 \midrule
Dequeue Max 
& 85.2 & 42.4 & 28.7 & 85.5 && 49.4 & 37.9 & 33.9 & 86.8 \\
FIFO 
& 86.8 & 44.2 & 31.2 & 85.2 && 59.2 & 47.6 & 40.5 & 83.1 \\
Dequeue Min & 88.2 & 55.9 & 44.8 & 83.2 \\
\bottomrule[1.2pt]
\end{tabular}
}
\caption{Ablation study on buffer update strategy. Swapping out the most confident sample with incoming ones (\ie, Dequeue Max) performs better than other alternatives.}
\label{tab:bufferType}
\end{table}
\begin{table}[]
\centering
\resizebox{\columnwidth}{!}{%
\begin{tabular}{@{}lllllcllll@{}}\toprule[1.2pt]
\multicolumn{1}{c}{\multirow{ 2}{*}{FPR@$\beta$TPR}} & \multicolumn{4}{c}{\textbf{CIFAR10}} & \phantom{a}& \multicolumn{4}{c}{\textbf{CIFAR100}} \\
\cmidrule{2-5} \cmidrule{7-10}
    & 98\%   & 95\%  & 92\% & AUC   && 98\%   & 95\%   & 90\% & AUC \\
 \midrule
Buffer = 0 
& 58.8 & 43.6 & 30.2 & 90.4 && 93.4 & 48.4 & 37.4 & 83.1 \\
Buffer = 5 
& 53.0 & 42.4 & 28.7 & 92.6 && 86.6 & 57.2 & 39.0 & 82.5 \\
Buffer = 10 
& 48.1 & 26.9 & 26.2 & 94.5 && 85.2 & 50.2 & 28.7 & 84.6 \\
Buffer = 20 
& 47.6 & 26.2 & 22.0 & 93.8 && 85.2 & 50.0 & 29.8 & 84.9 \\
Buffer = 32 
& 47.1 & 26.2 & 20.6 & 94.3 && 85.2 & 42.4 & 28.7 & 85.5 \\
Buffer = 48 & 46.1 & 24.3 & 23.2 & 93.9 && 85.2 & 42.1 & 27.6 & 85.5 \\
\bottomrule[1.2pt]
\end{tabular}
}
\caption{Illustration of the impact of the positive buffer. We report false positive rate results at high true positive rates for various buffer sizes.}
\label{tab:bufferSize}
\end{table}

\noindent\textbf{Melanoma.} 
We demonstrate the application of RankReg on imbalanced cancer classification using the Melanoma benchmark and show results in Table~\ref{tab:melanoma}. As we are the first to perform FPR vs. TPR study on such a large-scale dataset, there is a lack of comparison methods. Therefore, we provide results for all baselines as well as their combination with ALM by running experiments ourselves. 
It can be seen that, across all baselines, RankReg achieves state-of-the-art performance in the majority of metrics, with a minor setback on LDAM, where both ours and ALM achieve one best metric. 
These results verify the effectiveness of RankReg on a real-world dataset with critical positives.

\noindent\textbf{Discussion.} We attribute the performance lift obtained by RankReg to its more direct approach in ordering the critical positives ahead of the negatives, which is accomplished indirectly through margins in the previous state-of-the-art method ALM~\cite{alm}. The FPR at a given TPR depends only on how samples are ranked relative to each other, and not on the magnitude of the classification scores. Furthermore, the FPR at a high TPR is determined by the ranking of the least confident positives. Our regularizer places an increasing penalty on positive samples the lower they are ranked, which works to push up the scores of the hardest positives.

\subsection{Ablations and additional analyses}
In this section, we examine the impact of used components as well as provide additional evaluation to reveal our system's pros and cons. 
For these additional studies, unless otherwise indicated, we present results from using our approach with BCE as the base function.

\noindent\textbf{Rank penalty ablation}. Table \ref{tab:rankpenalty} shows an ablation study on different choices for the rank penalty in Eq.~\ref{eq:ell_reg}, including raw rank values $\bm{r}$, squared rank values $\bm{r}^{2}$, cubed rank values $\bm{r}^{3}$, and the exponential of rank values $e^{\bm{r}}$. Squared rank provides the best overall result while being simple.

\noindent\textbf{Impact of buffer maintenance strategy.} Our approach uses a buffer of critical positive samples to have meaningful ranking regularization signals at each batch of training. We evaluate the role of buffer by considering three kinds of maintenance strategies: (1) remove the most confident sample while adding new positive samples from a incoming batch (\ie, Dequeue Max), (2) first-in-first-out (\ie, FIFO), and (3) remove the least confident sample (\ie, Dequeue Min).
The results in Table~\ref{tab:bufferType} show that feeding sufficient amount of low-ranking positives to the model is useful, as evidenced by the increased performance across all metrics. 

\begin{figure}[t]
\begin{center}
   \includegraphics[width=.85\linewidth, trim={5 5 0 0}, clip]{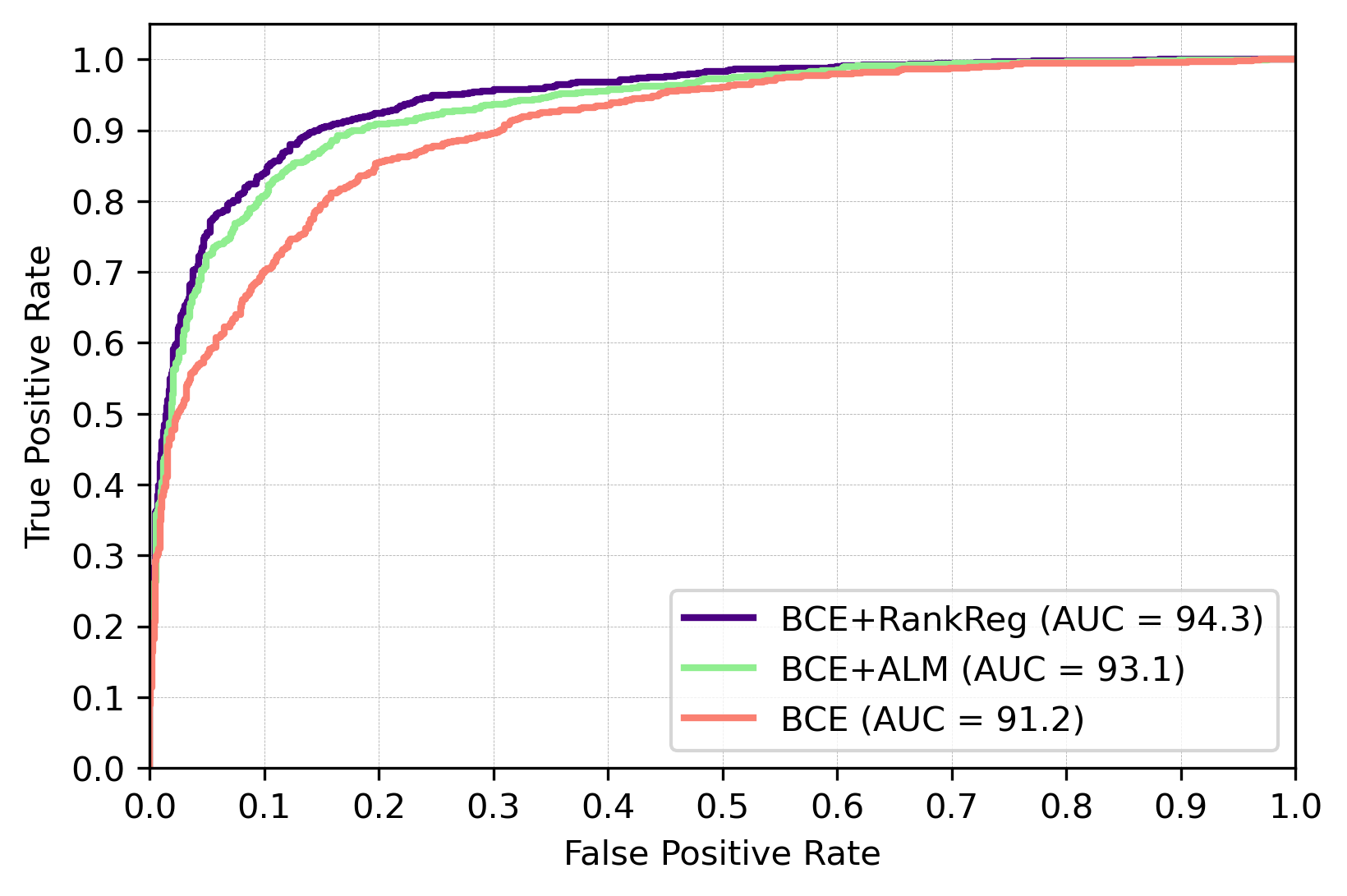}
\end{center}
\caption{Evalution of ROC results produced by our approach \textit{vs.} others on the CIFAR10 dataset (in 1:100 imbalance ratio).}
\label{fig:auc}
\end{figure}



\noindent\textbf{Impact of buffer size.} Throughout the empirical results in Sec.~\ref{sec:main_rst}, we use a buffer size of 32. 
Next, we ablate the size of the buffer by allowing the model to use more buffered positive samples during training. Table~\ref{tab:bufferSize} shows that the buffer plays an important role in our approach. Indeed, excluding the buffer component yields worse results; and performance (especially FPR@98TPR) improves quickly as buffer size increases. 32 seems to be an improving plateau.

\noindent\textbf{Visualization of ROCs.} To further estimate the effectiveness of our approach to reduce false positive rates at high true positive rates, we visualize our ROC curves as well as comparison methods, as shown in Figure~\ref{fig:auc}. The top two curves (\ie, ours and ALM) significantly surpass that of the BCE baseline on FPRs at earlier TPRs, \ie, starting from 30\% TPR and onward. Importantly, our approach performs on par with ALM up until $\sim$75\% TPR, and then consistently yields lower FPR values ever since to almost 100\% TPR.

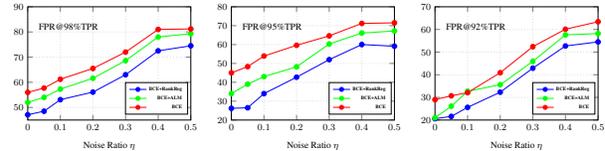
\begin{figure}[t]
    \centering
    \resizebox{0.15\textwidth}{!}{
\begin{tikzpicture}
\begin{axis}[
     title={FPR@98\%TPR},
     title style={at={(axis description cs:0.25,0.7)}, font=\footnotesize
    },
    legend style={
    nodes={scale=0.86, transform shape},
    cells={anchor=east},
    legend style={at={(0.79,0.05)},anchor=south,row sep=0.01pt}, font =\tiny} ,
    legend image post style={scale=0.86},
  xlabel={\footnotesize Noise Ratio $\eta$},
  xmin=0,xmax=0.5,
  ymin=45,ymax=90,
  ytick={50, 60, 70, 80, 90},
  xtick={0, 0.1, 0.2, 0.3, 0.4, 0.5},
  grid,
  grid style={dotted},
  x tick label style={font=\footnotesize},
  y tick label style={font=\footnotesize},
  width=6.5cm,
  height=5cm,
]

\addplot [line width=0.7pt, color=blue, style=solid, mark=*] coordinates {(0.0, 47.1) (0.05, 48.5) (0.1, 53.1) (0.2, 56.1) (0.3, 63) (0.4, 72.5) (0.5, 74.5)
};
\addplot [line width=0.7pt, color=green, style=solid, mark=*] coordinates {(0.0, 52.0) (0.05, 54.0) (0.1, 57.3) (0.2, 61.6) (0.3, 68.6) (0.4, 78.0) (0.5, 79.3)
};
\addplot [line width=0.7pt, color=red, style=solid, mark=*] coordinates {(0.0, 56.0) (0.05, 57.7) (0.1, 61.2) (0.2, 65.5) (0.3, 72) (0.4, 81) (0.5, 81.2)
};
\legend{ \textbf{BCE+RankReg}, \textbf{BCE+ALM}, \textbf{BCE}}

\end{axis}
\end{tikzpicture}
}
\resizebox{0.15\textwidth}{!}{
\begin{tikzpicture}
\begin{axis}[
    title={FPR@95\%TPR},
    title style={at={(axis description cs:0.25,0.7)}, font=\footnotesize
    },
    legend style={
    nodes={scale=0.86, transform shape},
    cells={anchor=east},
    legend style={at={(0.79,0.05)},anchor=south,row sep=0.01pt}, font =\tiny} ,
    legend image post style={scale=0.86},
  xlabel={\footnotesize Noise Ratio $\eta$},
  xmin=0,xmax=0.5,
  ymin=20,ymax=80,
  ytick={20, 30, 40, 50, 60, 70, 80},
  xtick={0, 0.1, 0.2, 0.3, 0.4, 0.5},
  grid,
  grid style={dotted},
  x tick label style={font=\footnotesize},
  y tick label style={font=\footnotesize},
  width=6.5cm,
  height=5cm,
]

\addplot [line width=0.7pt, color=blue, style=solid, mark=*] coordinates {(0.0, 26.2) (0.05, 26.5) (0.1, 34.0) (0.2, 42.7) (0.3, 52.0) (0.4, 60.0) (0.5, 59.1)
};
\addplot [line width=0.7pt, color=green, style=solid, mark=*] coordinates {(0.0, 34.0) (0.05, 39.0) (0.1, 43.0) (0.2, 48.2) (0.3, 60.3) (0.4, 66.1) (0.5, 67.2)
};
\addplot [line width=0.7pt, color=red, style=solid, mark=*] coordinates {(0.0, 45.) (0.05, 48.3) (0.1, 53.9) (0.2, 59.6) (0.3, 64.6) (0.4, 71.2) (0.5, 71.5)
};
\legend{ \textbf{BCE+RankReg}, \textbf{BCE+ALM}, \textbf{BCE}}
\end{axis}
\end{tikzpicture}
}
\resizebox{0.15\textwidth}{!}{
\begin{tikzpicture}
\begin{axis}[
    title={FPR@92\%TPR},
    title style={at={(axis description cs:0.25,0.7)}, font=\footnotesize
    },
    legend style={
    nodes={scale=0.86, transform shape},
    cells={anchor=east},
    legend style={at={(0.79,0.05)},anchor=south,row sep=0.01pt}, font =\tiny},
    legend image post style={scale=0.86},
  xlabel={\footnotesize Noise Ratio $\eta$},
  xmin=0,xmax=0.5,
  ymin=20,ymax=70,
  ytick={20, 30, 40, 50, 60, 70},
  xtick={0, 0.1, 0.2, 0.3, 0.4, 0.5},
  grid,
  grid style={dotted},
  x tick label style={font=\footnotesize},
  y tick label style={font=\footnotesize},
  width=6.5cm,
  height=5cm,
]

\addplot [line width=0.7pt, color=blue, style=solid, mark=*] coordinates {(0.0, 20.6) (0.05, 21.6) (0.1, 25.6) (0.2, 32.3) (0.3, 42.9) (0.4, 52.7) (0.5, 54.5)
};
\addplot [line width=0.7pt, color=green, style=solid, mark=*] coordinates {(0.0, 21.0) (0.05, 26.1) (0.1, 32.7) (0.2, 35.6) (0.3, 45.9) (0.4, 57.6) (0.5, 58.2)
};
\addplot [line width=0.7pt, color=red, style=solid, mark=*] coordinates {(0.0, 29.0) (0.05, 30.7) (0.1, 32.1) (0.2, 40.9) (0.3, 52.4) (0.4, 60.1) (0.5, 63.4)
};
\legend{ \textbf{BCE+RankReg}, \textbf{BCE+ALM}, \textbf{BCE}}
\end{axis}
\end{tikzpicture}
}
\caption{Label noise experiments using BCE as base loss on CIFAR10. We report FPR@\{98, 95, 92\}\%TPR (left to right) with varied noise ratios.}
\label{fig:my_label}
\end{figure}

\noindent\textbf{Robustness to label noise.} Real-world datasets often contain mislabeled data. To evaluate the robustness of our approach in the presence of label noise, we perform additional experiments in which we incrementally flip a proportion $\eta$ of training labels. Figure \ref{fig:my_label} shows how FPR@\{98, 95, 92\}\%TPR (left to right) degrade as a function of $\eta$ in the range of $[0, 0.5]$, using BCE as base loss. These results suggest that RankReg is as robust to label noise as the state-of-the-art approach \cite{alm}.


\noindent\textbf{1:200 imbalance ratio.} We also test our
model on more imbalanced situations, \eg, 1:200 imbalance ratio. To this end, we use the same data curation pipeline as introduced in Sec.~\ref{sec:dataset_eval} and build binary imbalanced CIFAR-10 and 100 datasets with a 1:200 imbalance ratio. We defer the full result tables to the supplementary material for space reasons. 
Looking at the averaged improvements (\ie, \textcolor{green!60!black}{Avg.}$ \Delta$) in the bottom row, our approach leads by a large margin.


\begin{table}[]
\centering
\resizebox{\columnwidth}{!}{%
\begin{tabular}{@{}llllclll@{}}\toprule[1.2pt]
\multicolumn{1}{c}{\multirow{ 2}{*}{Error@$\beta$\%TPR $\downarrow$}} & \multicolumn{3}{c}{\textbf{LT-CIFAR10} imb. 100} & \phantom{a} & \multicolumn{3}{c}{\textbf{LT-CIFAR10} imb. 200} \\
\cmidrule{2-4} \cmidrule{6-8}
    & 80\%   & 90\%  & Acc.   && 80\%   & 90\%  & Acc \\
 \midrule
CE & 29.8 & 34.7 & 70.4  && 37.8 & 42.4 & 64.0 \\
CE+ALM & 28.9 & 33.9 & 70.9 && 36.1& 39.9 & 65.1 \\
CE+RankReg & 26.7 & 29.3 & 71.6 && 36.7 & 37.8 & 65.0 \\
\bottomrule[1.2pt]
\end{tabular}
}
\caption{Multi-class experiments using long-tailed CIFAR-10. Baseline numbers are quoted from ALM~\cite{alm}.}
\label{tab:multiclass}
\end{table}
\noindent\textbf{Multi-class extension}. RankReg can be used in multi-class settings by ranking the critical samples higher than others based on the output probability for each class. Table \ref{tab:multiclass} shows additional results in the multi-class setting using long-tailed CIFAR-10 following the experiment protocol in \cite{alm}. We report the average error rate of other classes after setting thresholds for \{80, 90\}\%TPR on the critical class \cite{alm}. Our method performs better than \cite{alm} under the 1:100 imbalance ratio setting and comparably under the 1:200 setting.
\section{Conclusion}

The problem setting of critical rare positives has been surprisingly under-studied in the research community. This paper introduces a general method for inducing a neural network to prioritize the reduction of false positives when the operational context calls for a high true positive rate.
Motivated by the observation that the false positive rate at a high true positive rate is determined by how the least confident positives are ranked by the network, we formulated a ranking-based regularizer that places an increasing penalty on positive samples the lower they are ranked in a sorted list of the network's classification scores.
Experimental results show how our regularizer can be combined with a wide range of conventional losses and achieves state-of-the-art results in standard evaluations.
We hope that our findings will inspire broader interest in this important problem setting, as well as provide practitioners a simple yet effective method to train better neural network models for critical rare classes.
\newpage
\section*{A. Additional Results}

\begin{table}[h]
\small
\centering 
\resizebox{0.8\columnwidth}{!}{%
\begin{tabular}{l c c c c} 
\toprule 
& \multicolumn{4}{c}{Binary CIFAR10, imb. \textbf{1:200}} \\ %
\cmidrule(l){2-5} 
Methods & \vtop{\hbox{\strut FPR@ $\downarrow$}\hbox{\strut  98\%TPR}} & \vtop{\hbox{\strut FPR@ $\downarrow$}\hbox{\strut  95\%TPR}} &  \vtop{\hbox{\strut FPR@ $\downarrow$}\hbox{\strut  92\%TPR}} & AUC $\uparrow$ \\ 
\midrule 
BCE & 75.0 & 55.0 & 40.0 & 87.3 \\ 
+ALM  & 70.0 & 54.0  & 39.0 & 86.7\\ 
+RankReg   & \underline{\textbf{67.3}} & \underline{\textbf{52.7}}  & \underline{\textbf{37.8}} & \textbf{\cl{red!60!white}{89.6}} \\ 
\midrule
S-ML & 75.0 & 54.0  & \underline{\textbf{35.0}} & 87.4 \\ 
+ALM  & 72.0 & 52.0  & 39.0 & 87.9\\ 
+RankReg  & \underline{\textbf{65.8}} & \underline{\textbf{51.8}}  & 41.3 & \textbf{\cl{red!60!white}{88.2}} \\ 
\midrule
S-FL & 78.0 & 59.0  & 43.0 & 85.7 \\ 
+ALM  & 74.0 & 55.0  & 41.0 & 86.9\\ 
+RankReg   & \underline{\textbf{70.5}} & \underline{\textbf{46.5}} & \underline{\textbf{40.2}} & \textbf{\cl{red!60!white}{87.7}} \\ 
\midrule
A-ML & 74.0 & 56.0  & 39.0 & 87.4 \\ 
+ALM & 75.0 & 54.0  & 35.0 & 87.6\\ 
+RankReg  & \underline{\textbf{64.9}} & \underline{\textbf{50.4}}  & \underline{\textbf{38.1}} & \textbf{\cl{red!60!white}{89.2}} \\ 
\midrule
A-FL & 76.0 & 59.0  & 40.0 & 86.2 \\ 
+ALM & 78.0 & 57.0  & 37.0 & 87.0\\ 
+RankReg  & \underline{\textbf{68.6}} & \underline{\textbf{53.6}}  & \underline{\textbf{35.6}} & \textbf{\cl{red!60!white}{88.4}} \\ 
\midrule
CB-BCE & 87.0 & 74.0  & 61.0 & 78.0 \\ 
+ALM  & 85.0 & 69.0  & 53.0 & 80.0 \\ 
+RankReg   & \underline{\textbf{72.0}} & \underline{\textbf{54.8}} & \underline{\textbf{44.5}} & \textbf{\cl{red!60!white}{87.1}} \\ 
\midrule
W-BCE & 88.0 & 75.0  & 62.0 & 78.3 \\ 
+ALM  & 83.0 & 69.0  & 54.0 & 81.0 \\ 
+RankReg  & \underline{\textbf{66.3}} & \underline{\textbf{49.6}}  & \underline{\textbf{39.5}} & \textbf{\cl{red!60!white}{89.4}} \\ 
\midrule
LDAM & 78.0 & 63.0  & 45.0 & 86.4 \\ 
+ALM  & 73.0 & 61.0  & 43.0 & 85.6\\ 
+RankReg  & \underline{\textbf{65.8}} & \underline{\textbf{47.3}}  & \underline{\textbf{33.8}} & \textbf{\cl{red!60!white}{89.7}} \\ 
\midrule
\textcolor{green!60!black}{Avg.} $\Delta$ & \textcolor{Green}{8.2} & \textcolor{Green}{8.0} & \textcolor{Green}{3.2} & \textcolor{Green}{3.1} \\
\bottomrule 
\end{tabular}
}
\caption{
Comparison results for binary imbalanced CIFAR-10 showing FPRs at \{98\%, 95\%, 92\%\} TPRs. Baseline numbers are quoted from ALM~\cite{alm}. ``+ALM" and ``+RankReg" are shorthand for \textit{BaseLoss}+ALM and \textit{BaseLoss}+RankReg, respectively.
} 
\label{tab:cifar10_imb200} 
\end{table}
\newpage 
\vspace*{0.65em}
\begin{table}[h]
\small
\centering 
\resizebox{0.8\columnwidth}{!}{%
\begin{tabular}{l c c c c} 
\toprule 
& \multicolumn{4}{c}{Binary CIFAR100, imb. \textbf{1:200}} \\ %
\cmidrule(l){2-5} 
Methods & \vtop{\hbox{\strut FPR@ $\downarrow$}\hbox{\strut  98\%TPR}} & \vtop{\hbox{\strut FPR@ $\downarrow$}\hbox{\strut  95\%TPR}} &  \vtop{\hbox{\strut FPR@ $\downarrow$}\hbox{\strut  90\%TPR}} & AUC $\uparrow$ \\ 
\midrule 
BCE & 94.0 & 77.0 & 61.0 & 79.1 \\ 
+ALM  & 87.0 & \underline{\textbf{66.0}}  & 57.0 & 80.9\\ 
+RankReg   & \underline{\textbf{85.2}} & 66.8  & \underline{\textbf{43.0}} & \textbf{\cl{red!60!white}{83.1}} \\ 
\midrule
S-ML & 95.0 & 75.0  & 64.0 & 79.7 \\ 
+ALM  & 87.0 & 73.0  & 55.0 & 80.7\\ 
+RankReg  & \underline{\textbf{81.8}} & \underline{\textbf{62.8}}  & \underline{\textbf{48.4}} & \textbf{\cl{red!60!white}{82.4}} \\ 
\midrule
S-FL & 90.0 & 78.0  & 50.0 & 80.1 \\ 
+ALM  & 85.0 & 76.0  & 50.0 & 80.8\\ 
+RankReg   & \underline{\textbf{79.2}} & \underline{\textbf{65.8}} & \underline{\textbf{47.2}} & \textbf{\cl{red!60!white}{83.1}} \\ 
\midrule
A-ML & 95.0 & 75.0  & 66.0 & 79.8 \\ 
+ALM & 92.0 & 63.0  & 45.0 & 81.0\\ 
+RankReg  & \underline{\textbf{86.2}} & \underline{\textbf{59.0}}  & \underline{\textbf{43.0}} & \textbf{\cl{red!60!white}{83.4}} \\ 
\midrule
A-FL & 91.0 & 78.0  & 50.0 & 80.0 \\ 
+ALM & 88.0 & 76.0  & 46.0 & 80.7\\ 
+RankReg  & \underline{\textbf{81.2}} & \underline{\textbf{58.8}}  & \underline{\textbf{37.8}} & \textbf{\cl{red!60!white}{83.7}} \\ 
\midrule
CB-BCE & 93.0 & 78.0  & 51.0 & 78.7 \\ 
+ALM  & \underline{\textbf{85.0}} & 66.0  & 44.0 & 81.0 \\ 
+RankReg   & 85.0 & \underline{\textbf{64.0}} & \underline{\textbf{40.4}} & \textbf{\cl{red!60!white}{81.4}} \\ 
\midrule
W-BCE & 95.0 & 63.0  & 51.0 & 79.7 \\ 
+ALM  & 79.0 &62.0  & 44.0 & 81.3 \\ 
+RankReg  & \underline{\textbf{69.9}} &  \underline{\textbf{51.4}}  & \underline{\textbf{41.6}} & \textbf{\cl{red!60!white}{84.3}} \\ 
\midrule
LDAM & 80.0 & 67.0  & 45.0 & 82.1 \\ 
+ALM  & 84.0 & 61.0  & 46.0 & 81.5\\ 
+RankReg  & \underline{\textbf{70.2}} & \underline{\textbf{56.8}}  & \underline{\textbf{37.4}} & \textbf{\cl{red!60!white}{84.1}} \\ 
\midrule
\textcolor{green!60!black}{Avg.} $\Delta$ & \textcolor{Green}{5.5} & \textcolor{Green}{7.2} & \textcolor{Green}{6.0} & \textcolor{Green}{2.1} \\
\bottomrule 
\end{tabular}
}
\caption{
Comparison results for binary imbalanced CIFAR-100 showing FPRs at \{98\%, 95\%, 90\%\} TPRs.
Baseline numbers are quoted from ALM~\cite{alm}. ``+ALM" and ``+RankReg" are shorthand for \textit{BaseLoss}+ALM and \textit{BaseLoss}+RankReg, respectively.
} 
\label{tab:cifar100_imb200} 
\end{table}
\newpage

{\small
\bibliographystyle{ieee_fullname}
\bibliography{egbib}
}

\end{document}